\documentclass[10pt,twocolumn,letterpaper]{article}

\usepackage[pagenumbers]{cvpr} 

\usepackage[dvipsnames]{xcolor}
\usepackage{amsmath}
\usepackage{verbatim}
\usepackage{tabularray}
\usepackage{array}
\usepackage{rotating}
\usepackage{multirow}
\usepackage{ragged2e}
\usepackage{scalerel}
\usepackage{bbm}
\usepackage{graphicx}
\usepackage{wrapfig}
\usepackage{xspace}
\usepackage{fontawesome}
\usepackage{tikz}
\usepackage{booktabs} 
\usepackage{pifont}   
\usepackage{graphicx} 
\usepackage{wasysym}  
\usepackage{xcolor}
\usepackage{colortbl}
\usepackage{subcaption}
\usepackage{algorithm}
\usepackage{algorithmic}
\usepackage{wrapfig}

\usepackage{eccvabbrv}

\usetikzlibrary{arrows, positioning, shapes, calc}

\newcommand{\subheader}[1]{\vspace{0.25\baselineskip} \noindent \textbf{#1.}}

\newcommand{\pullup}{\vspace{-0.2\baselineskip}}
\newcommand{\pullupp}{\vspace{-0.4\baselineskip}}
\newcommand{\pulluppp}{\vspace{-0.8\baselineskip}}

\definecolor{cvprblue}{rgb}{0.21,0.49,0.74}
\usepackage[pagebackref,breaklinks,colorlinks,allcolors=cvprblue]{hyperref}

\def\paperID{*****} 
\def\confName{CVPR}
\def\confYear{2026}

\title{Autonomous and Self-Adaptive AI-Generated Image Identification Systems} 
\vspace{-1em}
\author{
	Aref Azizpour, Tai D. Nguyen, Matthew C. Stamm\\
	Drexel University\\
	Philadelphia, PA, USA\\
	{\tt\small {aa4639,tdn47,mcs382}@drexel.edu}
}
\vspace{-1em}

\begin{document}
\maketitle
\vspace{-2em}
\begin{abstract}

Detecting AI-generated images is increasingly challenging as new generative models rapidly emerge. Existing detectors perform well on known generators but often fail to recognize images from unseen ones and cannot be updated without human supervision. 
To address these challenges, we propose AS-AID, an Autonomous and Self-Adaptive AI-Generated Image Identification framework that autonomously detects AI-generated images, and integrates models of images from previously unseen generators without human intervention. AS-AID unifies three key capabilities: (1) an AI-Generated Image Identification subsystem. Given an unlabeled stream of images, this subsystem categorizes them into real, synthetic from known generators, and from unknown generators using an enhanced forensic embedding space; (2) a New Source Discovery subsystem that isolates high-purity clusters of new generators among images from unknown generators; and (3) an Autonomous Adaptation and Validation subsystem that updates the identification subsystem and validates each adaptation without human oversight according to a new generator discovery.
Extensive experiments demonstrate that AS-AID reliably discovers emerging generators, autonomously incorporates them into its model, and maintains high detection accuracy, 
 while existing detectors suffer substantial performance drop as new generators appear.


\end{abstract}\vspace{-1em}

\section{Introduction}

AI image generators can produce hyper-realistic images that can be misused for fraud, disinformation, and other harmful applications.
To counter this threat, numerous synthetic image detectors have been developed~\cite{PatchFor, CNNDet, DCTCNN, marra2018detection, DIF} that leverage forensic microstructures, \ie, invisible statistical traces left in an image by its source~\cite{ChangWIFS2019, corvi2023detection, marra2019gans, lukas2006digital}.
However, these detectors are trained on images from known generators, and new generators continually emerge with different microstructures, causing detectors to fail on images from these unseen sources~\cite{UFD, corvi2023detection, Cozzolino_2024_CVPR, vahdati2024beyond}.

Existing approaches to address this problem offer only partial solutions.
Transferable~\cite{NPR, UFD, LGrad, DE-Fake, RINE, DRCT} and zero-shot detectors~\cite{Aeroblade, Zed, fsd} aim to generalize to unseen generators, however their accuracy still degrades if these generators introduce microstructures unseen during training.
Open-set attribution methods~\cite{Fang_2023_BMVC, POSE, Abady} can flag images with out-of-distribution (OOD) microstructures,
but cannot discover when new generators emerge or update their detection model accordingly.
Continual learning methods~\cite{E3, marra2019incremental} allow detectors to be updated once a new generator is discovered, but a human must first identify the new generator, assemble training data, and
initiate retraining.

\begin{figure}[t] 
	\centering
	\includegraphics[width=0.85\linewidth]{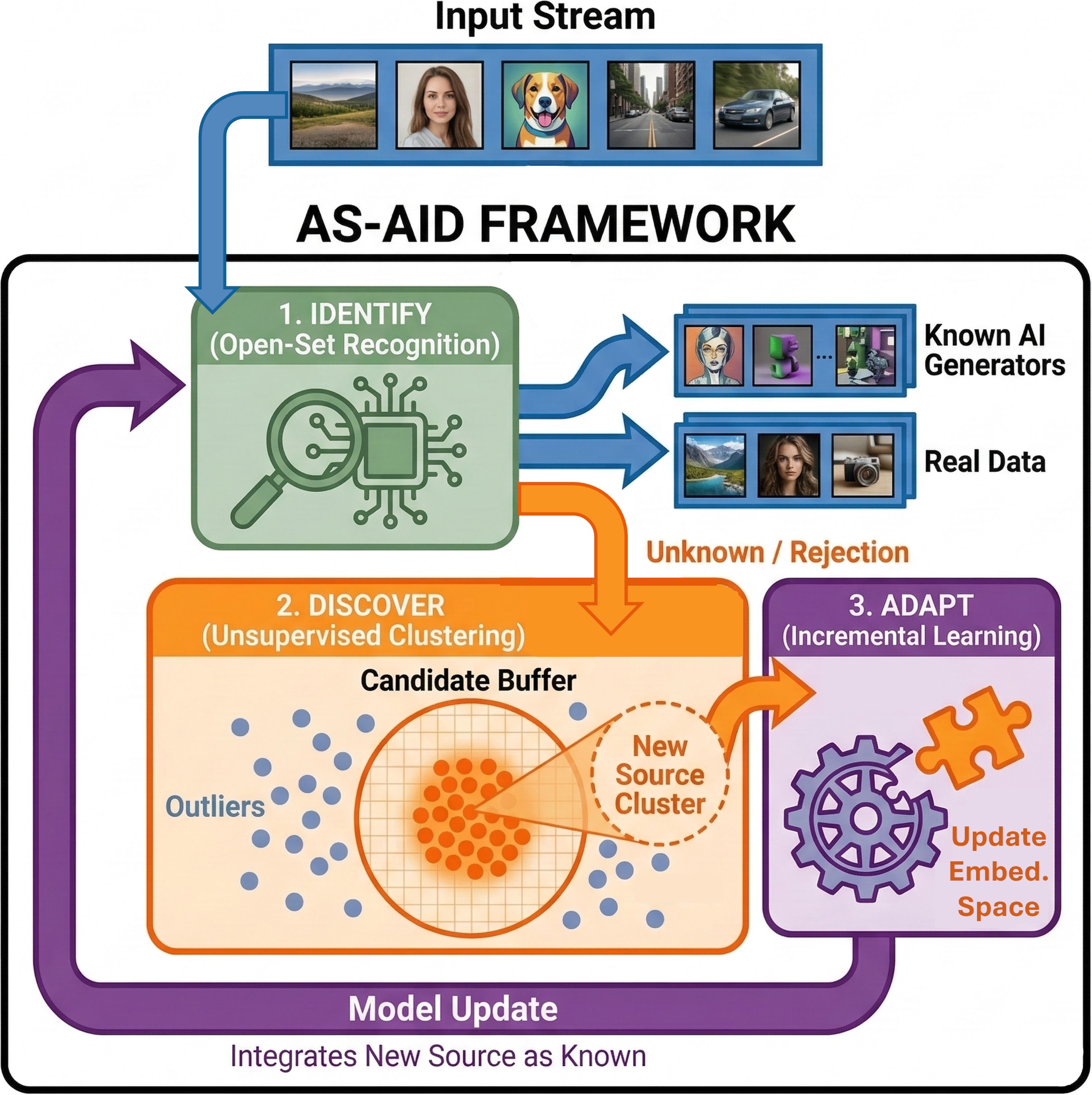}
	\caption{AS-AID system autonomously detects, discovers, and adapts to new AI image generators without human intervention.}
	\label{fig:front_page}
	\vspace{-1em}
\end{figure}

Currently, no existing method can autonomously detect the emergence of a new generator, construct a model of its microstructures, and integrate this model into the overall detection framework without human intervention.
This results in
several vulnerabilities: novel or undisclosed generators may go unnoticed, existing detectors may fail to identify images produced by these unseen models, and even once a new generator is discovered,
detector updates are slowed due to their reliance on human intervention and oversight.

To address these challenges, we introduce a new framework for synthetic media detection: \textbf{A}utonomous and \textbf{S}elf‑Adaptive \textbf{A}I‑Generated Image \textbf{Id}entification (AS‑AID).
This framework is able to
discover the emergence of a new
generator, then self-adapt by autonomously constructing a model of its microstructures and integrating this model into the detector - all without human supervision.
This is accomplished by utilizing three novel subsystems that enable our AS-AID system to maintain consistent synthetic image detection performance, even as new and previously undisclosed generators continually emerge.

\noindent
Our paper makes the following novel contributions:
\begin{itemize}

	\item 	A \textit{new synthetic media detection framework} that proposes  the first fully autonomous approach capable of discovering new generators, automatically updating its detection models after this discovery, and maintaining reliable detection performance without human intervention.

	\item A new \textit{AI-Generated Image Identification} subsystem that uses a novel embedding space to accurately
	identify synthetic
	images to known sources and reliably detect when an image originates from an unknown source.

	\item 	A  novel \textit{New Source Discovery} subsystem that accumulates images form unknown sources, identifies coherent image clusters with self-similar forensic traces, and
	test to determine if this corresponds to a new generator.

	\item A novel \textit{Autonomous Adaptation and Validation} subsystem that, upon discovery, updates both the embedding space and identification function, then validates that detection performance remains consistent.

	\item	\textit{Extensive experimental evaluation} showing that AS-AID reliably discovers newly emerging generators, maintains strong detection accuracy
	as generators continually emerge,
	and outperforms state-of-the-art detectors.
\end{itemize}

\section{Background and Related Work}
\label{sec:background}

Multiple approaches have been developed to enable the detection of images from new generative sources, including:

\subheader{Generalizable \& Zero-Shot Detection}
These methods extend detection beyond known generators by improving generalization to unseen sources~\cite{CNNDet, UFD, LGrad, NPR, c2p-clip, DRCT, RINE} or by modeling the distribution of real images independent of generator artifacts~\cite{Zed, fsd, Aeroblade}. Although initially transferable, their performance degrades as new generative architectures emerge (\eg, diffusion models replacing GANs)~\cite{corvi2023detection, chameleon, UFD}. These methods require manual updates when new sources appear, which requires human oversight.

\subheader{Source Attribution Methods}
Open-set attribution approaches~\cite{POSE, Fang_2023_BMVC} assign images to known sources~\cite{Repmix, ding2021does, wissmann2024whodunit} and flag unknown ones as out-of-distribution (OOD). However, like generalizable detectors, they gradually lose accuracy as new generative models emerge. Moreover, while they can flag OOD images, they cannot autonomously isolate coherent new sources among OOD data, limiting their scalability and requiring human intervention.

\begin{table}[t]
	\vspace{-3.4em}
	\centering
	\scriptsize
\caption{\small Comparison of Frameworks.}
	\label{tab:paradigm_comparison}
	\setlength{\tabcolsep}{1pt} 
	\begin{tabular}{l|c|c|c|c}
		\toprule
		\textbf{Framework} & \textbf{\begin{tabular}[c]{@{}c@{}}General-\\ization\end{tabular}} & \textbf{\begin{tabular}[c]{@{}c@{}}Source\\Attr.\end{tabular}} & \textbf{\begin{tabular}[c]{@{}c@{}}Adapt-\\ability\end{tabular}} & \textbf{\begin{tabular}[c]{@{}c@{}}Auto.\\Adapt.\end{tabular}} \\ \midrule
		Zero-Shot \& Transfer & \ding{51} & \ding{55} & \ding{55} & \ding{55} \\
		Open-Set Attribution & \RIGHTcircle & \ding{51} & \ding{55} & \ding{55} \\
		Continual Learning & \RIGHTcircle & \RIGHTcircle & \ding{51} & \ding{55} \\ \midrule
		\textbf{Ours} & \ding{51} & \ding{51} & \ding{51} & \textbf{\ding{51}} \\
		\bottomrule
	\end{tabular}
	\vspace{-1em}
\end{table}

\subheader{Continual Learning}
Continual learning techniques~\cite{E3, marra2019incremental} update models to accommodate new data while mitigating catastrophic forgetting~\cite{ewc, lwf, icarl}. Yet, they assume the availability of labeled data for each new source, relying on human-in-the-loop discovery and curation. This dependency makes them impractical in fast-evolving environments where new generators emerge continuously.

\section{Problem Formulation}
\label{sec:design_reqs}


\subsection{The Failure of Existing Approaches}
\label{sec:existing_failure}
Existing approaches lack the ability to autonomously adapt to the rapid emergence of new generative sources. 
This limitation is critical: these detectors cannot reliably recognize the distinct fingerprints of the novel generators without explicit retraining or manual intervention. As a result, their detection accuracy inevitably deteriorates over time.

\begin{figure}[t]
	\vspace{-2.2em}
	\centering
	\includegraphics[width=0.8\linewidth]{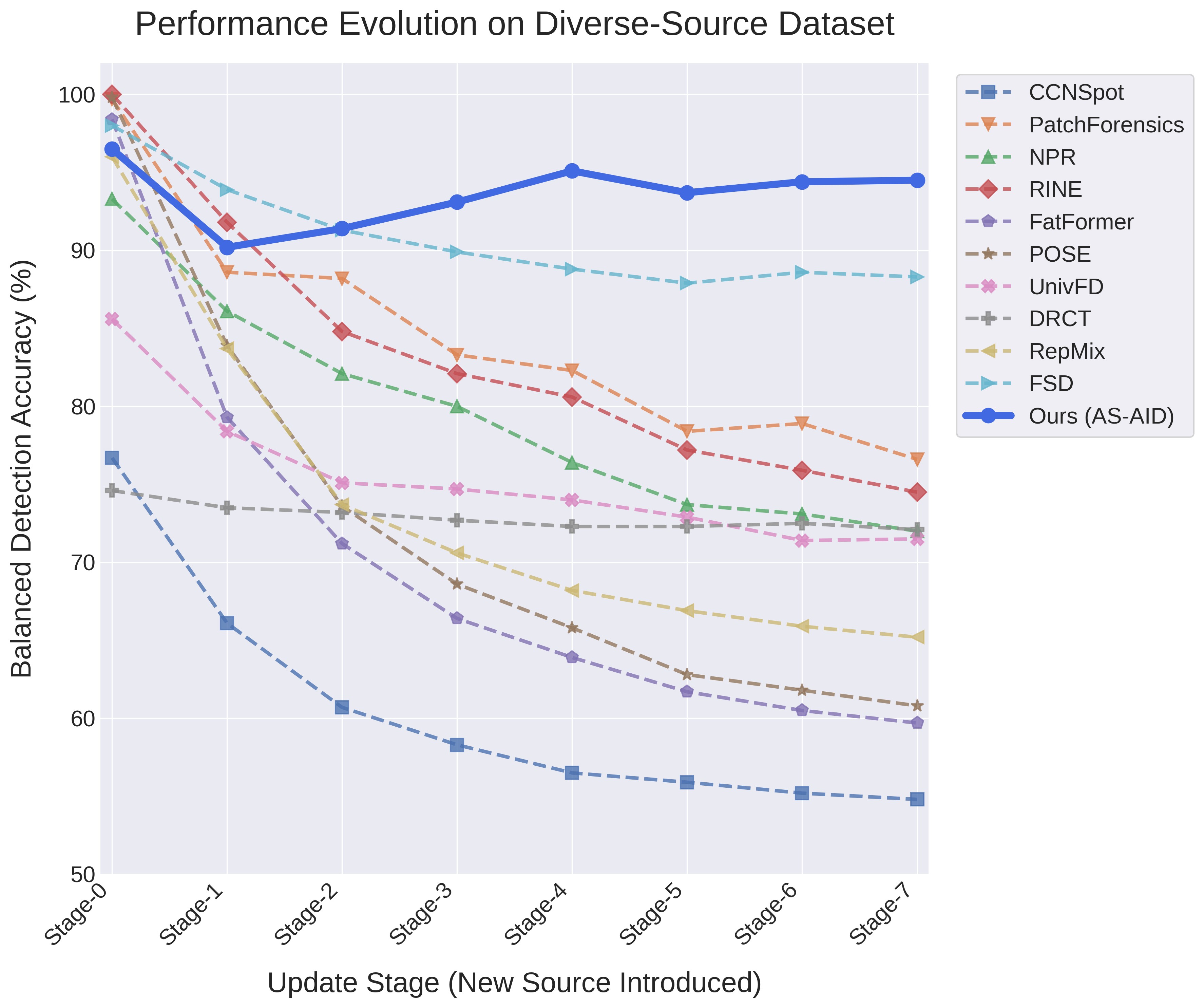}
	\vspace{-0.5em}
	\caption{Performance of methods as new generative sources emerge.}
	\label{fig:baseline_failure}
	\vspace{-1.5em}
\end{figure}

To demonstrate this limitation, we evaluated state‑of‑the‑art detectors in an evolving environment where new generative sources rapidly emerge. We sequentially introduced seven generators, including GAN‑based (\eg, GigaGAN~\cite{Gigagan}) and diffusion‑based (\eg, Stable Diffusion v1.5~\cite{stablediff}) models. Balanced detection accuracy at each stage is reported in Fig.~\ref{fig:baseline_failure} and Tab.~\ref{tab:main_results}. These results show that existing methods suffer substantial performance drop as new sources appear. Top methods such as POSE~\cite{POSE} and RINE~\cite{RINE} drop 25–39 points, highlighting the need for a framework that sustains strong detection performance under continual source emergence.

%

\subsection{Autonomous Self-Adaptive Identification}

The previous experiment highlights a critical limitation of existing synthetic image detectors: transferability alone is insufficient to maintain detection performance as new generative models continuously emerge. In practice, maintaining detector performance requires significant human oversight: identifying new generators, collecting training data, and manually updating models.

To overcome this limitation, we introduce a new class of systems termed \textbf{A}utonomous and \textbf{S}elf-Adaptive \textbf{A}I-Generated Image \textbf{Id}entification (\textbf{AS-AID}) systems. These systems are designed to maintain reliable detection performance in continuously evolving environments without requiring human intervention.

To achieve this goal, an AS-AID system must be capable of performing the following tasks autonomously:
(1)	\textbf{Discover the emergence of a new generator} whose images contain previously unseen forensic microstructures.
(2)	\textbf{Self-curate a set of training data} originating from this generator that can be used to update the detection model.
(3)	\textbf{Autonomously update the synthetic image identification system} using the newly curated training data and {validate that this update does not degrade detection performance}.

At present, no existing synthetic image detection system can accomplish these objectives \textit{without human involvement}. To enable autonomous self-adaptation, we argue that an AS-AID system must incorporate the following key subsystems.

\subheader{Open-Set Synthetic Image Identification} Binary detectors that simply distinguish between real and synthetic images are insufficient for discovering new generators. Instead, the system must determine whether an image is synthetic, attribute it to a known generator when possible, and identify when it cannot be confidently associated with any known source. Without this open-set capability, the system cannot detect the emergence of previously unseen generators.

\subheader{Autonomous Discovery of New Generators} An image that cannot be attributed to a known generator does not necessarily indicate a new source - it may simply be an anomalous, out-of-distribution sample from a known one. To reliably discover new generators, the system must accumulate unattributed images and identify a coherent subset with self-similar forensic microstructures. Such a subset signals a new generator and provides data for modeling it. No current detection system has this capability.

\subheader{Autonomous Adaptation with Self-Validation} Simply retraining on self-curated data from a discovered generator is insufficient - incorrectly labeled images cause data contamination, and naive updates may trigger catastrophic forgetting. Updates must therefore use a continual learning mechanism that incorporates new data while preserving prior knowledge. Since these updates occur without human oversight, the system must also validate that each update does not degrade overall performance.



\section{Proposed Method}
\label{sec:method}

\begin{figure}[t]
	\vspace{-2.4em}
	\centering
	\resizebox{0.9\linewidth}{!}{%
		\scalebox{1}[0.8]{%
			\includegraphics[width=\linewidth]{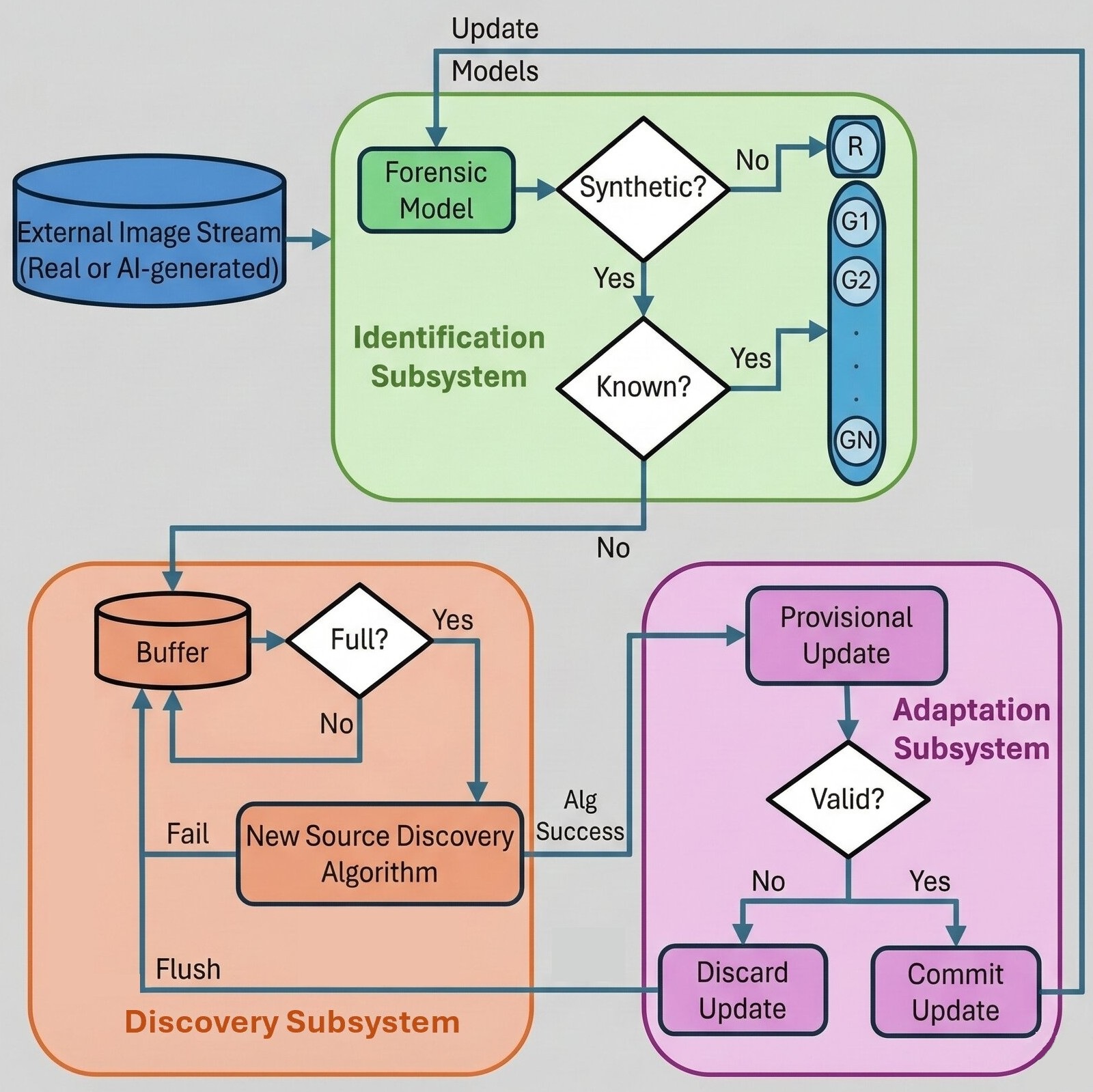}%
		}%
	}
	\caption{Flowchart of our AS-AID framework.}
	\label{fig:system_diagram}
	\vspace{-1.5em}
\end{figure}

Here, we propose a new AS-AID system combining three different subsystems, each addressing one of the challenges outlined in Sec.~\ref{sec:design_reqs}. 
A high-level overview is visualized in Fig.~\ref{fig:front_page} and a more detailed flow chart is shown in Fig.~\ref{fig:system_diagram}. 


\subsection{System-Level Overview}

Our system operates on a stream of images provided to it. The stream may include real images and synthetic images from known or unknown generators. The system is then tasked to detect synthetic images and identify if synthetic images are generated from a known or an unknown generator. Images that cannot be confidently attributed to any known source (stored in a set $\mathcal{S}$) are autonomously routed to an unknown-generator buffer $\mathcal{B}$ with fixed size. 

Once this buffer is full, our New Source Discovery subsystem analyzes $\mathcal{B}$ to isolate a subset of images likely originating from a previously unknown generator.
When a new generator is discovered, our Autonomous Adaptation and Validation subsystem builds a model of the new source and, before committing the update, performs a validation check to ensure system performance is not degraded.

\subheader{Forensic Microstructures}
Prior research in multimedia forensics shows image creation processes introduce subtle, invisible forensic microstructures~\cite{ChangWIFS2019, corvi2023detection, marra2019gans}. These microstructures correspond to statistical inter-pixel dependencies arising from the source's internal computation pipeline (\eg, camera or AI-generator)~\cite{lukas2006digital, marra2019gans, ChangWIFS2019, zhao2016computationally}. Like prior research~\cite{NPR, CNNDet, RINE, PatchFor}, we leverage these microstructures to perform synthetic image detection.

\subheader{Capturing Microstructures}
We capture these microstructures using an embedding $\phi = h(x)$, where $h(\cdot)$ is an embedding function. Traditionally, $h(\cdot)$ has been trained in a supervised fashion to discriminate among a set of known real and generative sources~\cite{NPR, CNNDet, PatchFor, fatformer}. While effective initially, recent studies show that these embeddings tend to discard microstructures not directly useful for discriminating the training sources. As a result, they exhibit reduced performance on previously unseen generators~\cite{chameleon, corvi2023detection, UFD}.

To overcome this, we adopt the self-supervised method from Nguyen \etal~\cite{fsd} as our choice of $h(\cdot)$. The resulting embeddings, $\phi = h(x)$, are known as {Forensic Self-Descriptions (FSDs)}: compact representations of \emph{all} forensic traces in an image, learned without supervision tied to any generator. This preserves the microstructural information needed for robust detection and critically enables reliable discovery and characterization of unseen generators.

\subsection{AI-Generated Image Identification}
\label{sec:identification_module}
%

At the heart of our AS-AID system lies a novel AI-Generated Image Identification subsystem. This subsystem functions by projecting an image's initial embedding $\phi = h(x)$ into an \textit{Enhanced Forensic Embedding Space} designed to enforce source separability. Within this optimized space, the system subsequently assigns a detection and attribution decision.

\begin{figure}[t]
	\vspace{-2.4em}
	\setlength{\abovecaptionskip}{4pt}
	\centering
	\begin{subfigure}{0.4\linewidth}
		\centering
		\includegraphics[width=\linewidth]{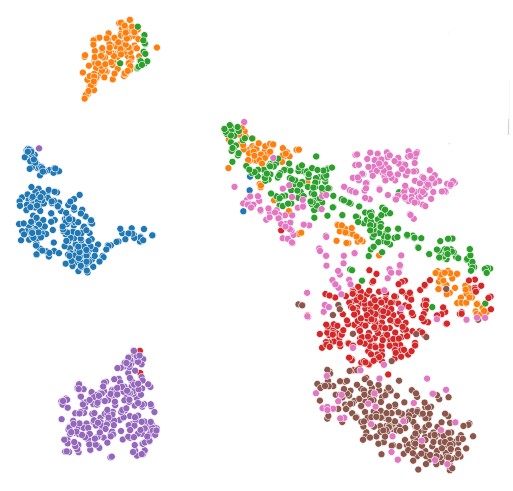}
		\caption{}
		\pullup
		\label{fig:embed_before}
	\end{subfigure}
	\hfill\vrule\hfill
	\begin{subfigure}{0.4\linewidth}
		\centering
		\includegraphics[width=\linewidth]{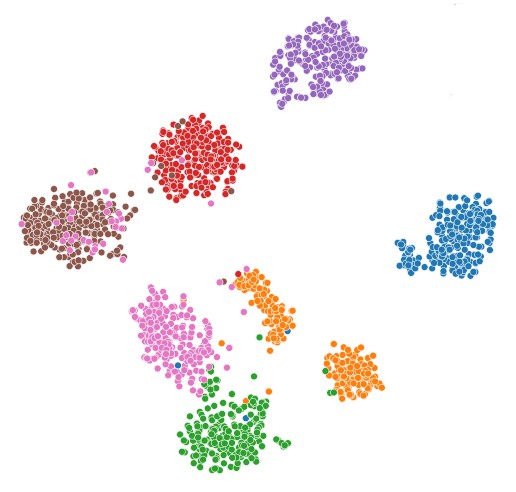}
		\caption{}
		\pullup
		\label{fig:embed_after}
	\end{subfigure}
	
	\caption{2D t-SNE plot of the embedding space before (a) and after (b) applying the enhanced source separation projection}
	\label{fig:embed_before_after_projection}
	\vspace{-1em}
\end{figure}

A challenge, however, is that while the initial self-supervised embeddings, $\phi$, are effective for zero-shot detection, they are not specifically optimized for source-level separability. Hence, embeddings from different sources may overlap, as shown in Fig.~\ref{fig:embed_before}. This low separability makes it difficult to accurately attribute an image to a known source and to determine whether it originates from a new, unknown generator.

\subheader{Enhanced Forensic Embedding Space}
\label{sec:enhanced_embedding}
To address the above limitations, we propose utilizing a projection network $f$. This network projects $\phi$, into an Enhanced Forensic Embedding Space, $\Psi$. Network $f$ transforms $\phi$ so that representations from different generators form distinct, well‑separated clusters, while still preserving forensic microstructures essential for generalizable detection.

The projection network {$f$} is designed to achieve two complementary objectives. The first is (i) to maximize source separability in $\Psi$ to make sure each image can be confidently and accurately attributed to its source. 
The second objective is (ii) to maintain strong zero-shot detection capability by preserving the rich, self-supervised forensic traces captured by the original $\phi$ space.

To do this, we propose a novel Separation-Preservation loss function to train $f$. This loss function is defined as:
\pullup
\begin{equation}
	\label{eq:enhance_emb_loss}
	\mathcal{L}_{\mathrm{e}} = \mathcal{L}_{\mathrm{s}} + \lambda\,\mathcal{L}_{\mathrm{p}},
\pullup
\end{equation}
where $\mathcal{L}_{\mathrm{s}}$ and $\mathcal{L}_{\mathrm{p}}$ are two specialized terms defined below, and $\lambda$ is a hyperparameter balancing the two terms. 

\begin{enumerate}
	\item \textbf{Separation Term ($\mathcal{L}_{\mathrm{s}}$):} To achieve objective (i), we enforce source separability by employing a triplet loss~\cite{facenet} on the projected embeddings $\psi$:
	\pullup
	\begin{equation}
	\label{eq:sep_loss}
	\pullup
	\hspace{-1em}
			\mathcal{L}_{\mathrm{s}}
			= 
			\sum\limits_{(a,p,n) \in \mathcal{T}}
			\Bigl[
			\| f(\phi^{a}) - f(\phi^{p}) \|
			-
			\| f(\phi^{a}) - f(\phi^{n}) \|
			+
			m
			\Bigr]_{+},
	\end{equation}
	where $(a,p,n)$ denotes an anchor $\phi^a$, a positive $\phi^p$ from the same source, and a negative $\phi^n$ from a different source, and $m$ is a margin hyperparameter.
	
	\item \textbf{Preservation Term ($\mathcal{L}_{\mathrm{p}}$):} To achieve objective (ii), we utilize a decoder network $g$, jointly trained with $f$,
	that attempts to reconstruct $\phi$ from the enhanced forensic embedding $f(\phi)$. This ensures that $f(\phi)$ retains the necessary forensic information encoded in $\phi$:
	\begin{equation}
		\smash{\mathcal{L}_{\mathrm{p}} = \| \phi - g(f(\phi)) \|_2^2.}
	\end{equation}
\end{enumerate}

As illustrated in Fig.~\ref{fig:embed_before_after_projection}, after training $f$  with $\mathcal{L}_e$, it transforms the original embeddings into an enhanced embedding space, $\Psi$, with improved separability, enabling reliable attribution and facilitating new-source discovery, while preserving strong detection capabilities.

%
%

\noindent \textbf{Forensic Identification.} We model enhanced embedding distributions for each known source $s \in \mathcal{S}$ using Gaussian Mixture Models (GMMs). We define an identification function $\alpha(\cdot)$ that assigns an image to the source with the highest GMM likelihood if it exceeds a confidence threshold $\tau_{id}$; otherwise flagged as unknown ($s_u$). By assigining a specific GMM for the real class ($s_{real}$), we unify detection ($\alpha(\cdot) \neq s_{real}$), attribution, and unknown identification.

\subsection{New Source Discovery}
\label{sec:discovery}

\begin{figure}[t]
	\vspace{-4.7em}
	\begin{minipage}{\linewidth}
		\begin{algorithm}[H]
			\caption{New Source Discovery Algorithm}
			\label{alg:discovery}
			\begin{algorithmic}[1]
				\STATE \textbf{Input:} Unknown embedding buffer $\mathcal{B}$, thresholds $\tau_s, \tau_c$
				\STATE \textbf{Output:} New source candidate cluster $\mathcal{C}^*$ or $\emptyset$
				\STATE Cluster embeddings in $\mathcal{B}$, store results in $\mathcal{K}$
				\STATE Sort $\mathcal{K}$ by cluster size (descending)
				\FOR{$\mathcal{C}_l$ \textbf{in} $\mathcal{K}$}
				\STATE Compute sufficiency $\gamma(\mathcal{C}_l)$ and cohesion $\eta(\mathcal{C}_l)$
				\IF{$\gamma(\mathcal{C}_l) > \tau_s$ \AND $\eta(\mathcal{C}_l) > \tau_c$}
				\STATE \textbf{return} $\mathcal{C}^* \gets \mathcal{C}_l$ \hfill // New source
				\ENDIF
				\ENDFOR
				\STATE \textbf{return} $\emptyset$ \hfill // No new source
			\end{algorithmic}
		\end{algorithm}
	\end{minipage}
	\vspace{-0.5em}
\end{figure}

Images that cannot be attributed to a known source do not necessarily originate from a new generator; they often include real outliers or hard-to-attribute images from known generators. As a result, simply updating on unknowns leads to model corruption. To address this, we propose the New Source Discovery Algorithm (NSDA)  to isolate a subset of self-similar, internally consistent embeddings in the buffer that reliably represent a new generator. A pseudo-code is provided in Alg.~\ref{alg:discovery}.

We accumulate unknown images in a buffer $\mathcal{B} = \{\psi_k | \alpha(x_k) = s_u \}$ of fixed size $b$. Once full, the algorithm groups the embeddings in $\mathcal{B}$ using DBSCAN~\cite{dbscan}, which is robust to noise and does not require a predefined number of clusters. We then analyze the resulting clusters $\mathcal{K} = \{\mathcal{C}_1, \dots, \mathcal{C}_L\}$, sorted by size, against two criteria to isolate a coherent new source:

\subheader{1. Sufficiency Criterion ($\gamma > \tau_s$)}
To justify a discovery decision, there must be sufficient evidence that the cluster represents a significant phenomenon rather than a transient anomaly. We define the sufficiency score as the ratio of the cluster size to the buffer size: $\gamma(\mathcal{C}_l) = |\mathcal{C}_l|/|\mathcal{B}|$. We require $\gamma(\mathcal{C}_l) > \tau_s$.

\subheader{2. Cohesion Criterion ($\eta > \tau_c$)}
A cluster that is sufficiently large may still lack internal consistency, potentially merging multiple distinct outlier groups into one. By contrast, images from a single new generator should form a dense cluster, tightly grouped to reflect their common origin and clearly separated from other clusters. To ensure this, we compute the average Silhouette Score~\cite{Silhouette} for the cluster: $\eta(\mathcal{C}_l) = |\mathcal{C}_l|^{-1} \sum_{\psi \in \mathcal{C}_l} s(\psi)$. We require $\eta(\mathcal{C}_l) > \tau_c$.

\subheader{Algorithm Outcome} The first cluster satisfying both criteria is returned as the new source candidate $\mathcal{C}^*$ to the next sub-system. If no cluster qualifies, the buffer $\mathcal{B}$ is flushed and accumulation restarts. This ensures that the system adapts \textit{only} when strong evidence of a coherent, distinct new source is present.

\subsection{Autonomous Adaptation and Validation}
\label{sec:validation}

Upon discovery of a new source, the system must adapt its embedding space and identification function. While standard approaches rely on human oversight to prevent data contamination, we propose an \textit{Autonomous Adaptation and Validation} subsystem. This module executes a provisional update using the discovered data, then validates that system's performance does not significantly degrade before committing the change.

\noindent \textbf{Autonomous Provisional Update.} 
At stage $\ell$, the system maintains known sources $\mathcal{S}^{(\ell)}$ and datasets $\mathcal{D}_T^{(\ell)}, \mathcal{D}_V^{(\ell)}$. Upon discovering candidate cluster $\mathcal{C}^*$, we partition it into training/validation splits $\mathcal{C}^*_T, \mathcal{C}^*_V$ and augment the system state: $\mathcal{S}^{(\ell+1)} \leftarrow \mathcal{S}^{(\ell)} \cup \{ s_{n+1} \}$ and $\mathcal{D}_T^{(\ell+1)} \leftarrow \mathcal{D}_T^{(\ell)} \cup \mathcal{C}^*_T$.
Adaptation then proceeds in two steps:
(1) \textbf{Embedding Refinement:} We refine the projection network $f_\theta$ using the expanded training set $\mathcal{D}_T^{(\ell+1)}$ to minimize $\mathcal{L}_{\mathrm{e}}$ (Eq.~\ref{eq:enhance_emb_loss}). This ensures the embedding space captures the new source's microstructures while preserving the the original forensic traces captured by the embedding space.
(2) \textbf{Identification Update:} We generate updated embeddings for all known sources using $f^{(\ell+1)}$ and re-fit the source-specific GMMs to update $\alpha(\cdot)$.

\noindent \textbf{Autonomous Validation.} 
Since discovery is unsupervised, $\mathcal{C}^*$ may mistakenly contain images that do not belong to the new source. Performing adaptation with this data may introduce data contamination~\cite{Mentornet}. To prevent this, we validate the provisional model on $\mathcal{D}_V^{(\ell+1)} \cup \mathcal{C}^*_V$ against two criteria before committing:
\begin{enumerate}
	\item \textbf{Detection Stability ($\nu_1$):} The drop in overall detection accuracy ($\nu_1$) must be minimal ($\nu_1 \le \epsilon_1$), ensuring images from old generators remain detectable.
	\item \textbf{Modeling Quality ($\nu_2$):} The attribution accuracy for the new source ($\nu_2$) must be high ($\nu_2 \ge 1 - \epsilon_2$), ensuring a distinct model has been formed.
\end{enumerate}
The update is accepted if only both conditions are satisfied and discarded if either condition fails, reverting the system to state $\ell$.


\section{Experiments and Results}
\label{sec:experiments}

\subheader{Datasets}
\label{sec:benchmarks}
We evaluate all methods on a common set of evaluation tasks, summarized in Tab.~\ref{tab:datasets}.  \noindent This includes a consistent initialization set for training and two evaluation sets, the GenImage Dataset (GID) and the Diverse-Source Dataset (DSD) to assess sequential adaptation. GID is utilized as it serves as a standard benchmark in recent literature. To rigorously evaluate a "worst-case" scenario, DSD is explicitly constructed with generators exhibiting substantially different forensic traces. This design prevents the artificially inflated performance metrics that arise when testing on highly similar model variants, such as fine-tunes of the same base architecture. Further details on the exact splits are provided in the supplementary material.


\begin{table}[t]
	\centering
	\caption{Datasets used in the paper. We categorize our data into initial training, and two evaluation datasets.}
	\label{tab:datasets}
	\small
	\begin{tabular}{@{}ll@{}}
		\toprule
		\textbf{Purpose} & \textbf{Source(s) / Generators} \\
		\midrule
		\textbf{Initial Train} & ImageNet (Real)~\cite{imagenet} + Stable Diffusion v1.4~\cite{stablediff} \\
		\noalign{\smallskip}
		\midrule
		\noalign{\smallskip}
		\textbf{Eval 1 (GID)} & \textbf{GenImage}~\cite{GenImage} (ADM~\cite{adm}, BigGAN~\cite{biggan}, Glide~\cite{glide}, \\
		& Midjourney~\cite{Midjourney}, SD v1.5~\cite{stablediff}, VQDM~\cite{vqdm}, Wukong) \\
		\noalign{\smallskip}
		\textbf{Eval 2 (DSD)} & \textbf{Diverse-Source}~\cite{Fang_2023_BMVC, Synthbuster, GenImage} (GigaGAN~\cite{Gigagan}, MJv5-6~\cite{Midjourney}, \\
		& VQDM~\cite{vqdm}, Flux~\cite{Flux}, DALL·E 3~\cite{dalle3}, Firefly~\cite{Firefly}) \\
		\bottomrule
	\end{tabular}
	\vspace{-1em}
\end{table}

\begin{table*}[t]
	\centering
	\scriptsize

	\caption{Overall \textbf{balanced detection accuracy} (\%) on two datasets, Diverse-Source Dataset (left), and GenImage Dataset (right), as new generative sources emerge. The \textbf{Drop} column is the difference in performance between first and last stage.}
	\label{tab:main_results}
	\setlength{\tabcolsep}{3.0pt}

	\begin{minipage}{.49\linewidth}
		\centering
		\begin{tabular}{l cccccccc | c}
			\toprule
			 & \multicolumn{8}{c}{\textbf{Update Stage}-$(\ell)$} & \\
			\cmidrule(lr){2-9}
			\textbf{Method} &\textbf{ S-0} & \textbf{S-1} & \textbf{S-2} & \textbf{S-3} & \textbf{S-4} & \textbf{S-5} & \textbf{S-6} & \textbf{S-7} & \textbf{Drop} \\
			\midrule
			CCNSpot~\cite{CNNDet}  & 76.7 & 66.1 & 60.7 & 58.3 & 56.5 & 55.9 & 55.2 & 54.8 & 21.9 \\
			PatchFor.~\cite{PatchFor} & 99.7 & 88.6 & 88.2 & 83.3 & 82.3 & 78.4 & 78.9 & 76.6 & 23.1 \\
			NPR~\cite{NPR}         & 93.3 & 86.1 & 82.1 & 80.0 & 76.4 & 73.7 & 73.1 & 72.0 & 21.3 \\
			RINE~\cite{RINE}       & \textbf{100.0} & 91.8 & 84.8 & 82.1 & 80.6 & 77.2 & 75.9 & 74.5 & 25.5 \\
			FatFormer~\cite{fatformer} & 98.4 & 79.3 & 71.2 & 66.4 & 63.9 & 61.7 & 60.5 & 59.7 & 38.7 \\
			POSE~\cite{POSE}       & 99.8 & 83.9 & 73.6 & 68.6 & 65.8 & 62.8 & 61.8 & 60.8 & 39.0 \\
			UnivFD~\cite{UFD}      & 85.6 & 78.4 & 75.1 & 74.7 & 74.0 & 72.9 & 71.4 & 71.5 & 14.1 \\
			DRCT~\cite{DRCT}       & 74.6 & 73.5 & 73.2 & 72.7 & 72.3 & 72.3 & 72.5 & 72.1 & \textbf{2.6} \\
			RepMix~\cite{Repmix}   & 96.0 & 83.7 & 73.7 & 70.6 & 68.2 & 66.9 & 65.9 & 65.2 & 30.8 \\
			FSD~\cite{fsd}         & 98.0 & \textbf{93.9} & 91.3 & 89.9 & 88.8 & 87.9 & 88.6 & 88.3 & 9.7 \\
			\cmidrule(lr){1-10}
			\cellcolor{gray!15}\textbf{Ours} & \cellcolor{gray!15}98.5 & \cellcolor{gray!15}90.2 & \cellcolor{gray!15}\textbf{91.4} & \cellcolor{gray!15}\textbf{93.1} & \cellcolor{gray!15}\textbf{95.1} & \cellcolor{gray!15}\textbf{93.7} & \cellcolor{gray!15}\textbf{94.4} & \cellcolor{gray!15}\textbf{94.5} & \cellcolor{gray!15} {4.0} \\
			\bottomrule
		\end{tabular}
	\end{minipage}
	\hfill
	\begin{minipage}{.49\linewidth}
		\centering
		\begin{tabular}{l cccccccc | c}
			\toprule
			& \multicolumn{8}{c}{\textbf{Update Stage}-$(\ell)$} & \\
			\cmidrule(lr){2-9}
			\textbf{Method} &\textbf{ S-0} & \textbf{S-1} & \textbf{S-2} & \textbf{S-3} & \textbf{S-4} & \textbf{S-5} & \textbf{S-6} & \textbf{S-7} & \textbf{Drop} \\
			\midrule
			CCNSpot~\cite{CNNDet}  & 76.7 & 65.5 & 62.6 & 60.7 & 59.1 & 58.6 & 59.1 & 59.5 & 17.2 \\
			PatchFor.~\cite{PatchFor} & 99.7 & 75.0 & 76.3 & 74.1 & 71.1 & 75.6 & 79.1 & 78.3 & 21.6 \\
			NPR~\cite{NPR}         & 93.3 & 83.0 & 81.6 & 79.5 & 77.7 & 77.0 & 77.4 & 77.8 & 15.5 \\
			RINE~\cite{RINE}       & \textbf{100.0} & 84.9 & 81.0 & 79.5 & 78.0 & 76.7 & 77.3 & 77.2 & 22.7 \\
			FatFormer~\cite{fatformer} & 98.4 & 78.1 & 74.8 & 72.5 & 70.9 & 69.9 & 70.6 & 71.2 & 27.2 \\
			POSE~\cite{POSE}       & 99.8 & 81.4 & 78.0 & 76.4 & 75.4 & 74.3 & 74.7 & 75.1 & 24.6 \\
			UnivFD~\cite{UFD}      & 85.6 & 78.4 & 76.2 & 74.6 & 73.4 & 72.8 & 72.6 & 73.1 & 12.5 \\
			DRCT~\cite{DRCT}       & 74.6 & 72.4 & 71.7 & 71.8 & 71.8 & 71.6 & 71.5 & 71.5 & \textbf{3.1} \\
			RepMix~\cite{Repmix}   & 96.0 & 78.4 & 74.8 & 71.6 & 69.4 & 68.3 & 69.2 & 69.8 & 26.1 \\
			FSD~\cite{fsd}         & 98.0 & 93.2 & 90.7 & 89.0 & 86.8 & 86.2 & 88.0 & 87.1 & 11.9 \\
			\cmidrule(lr){1-10}
			\cellcolor{gray!15}\textbf{Ours} & \cellcolor{gray!15}98.5 & \cellcolor{gray!15}\textbf{95.8} & \cellcolor{gray!15}\textbf{95.5} & \cellcolor{gray!15}\textbf{95.8} & \cellcolor{gray!15}\textbf{95.4} & \cellcolor{gray!15}\textbf{94.9} & \cellcolor{gray!15}\textbf{95.6} & \cellcolor{gray!15}\textbf{95.4} & \cellcolor{gray!15}\textbf{3.1} \\
			\bottomrule
		\end{tabular}
	\end{minipage}
\end{table*}

\subheader{Competing Methods}
We compare our proposed method against a set of detection and source attribution methods. These include zero-shot and generalizable detectors (CCNSpot~\cite{CNNDet}, UnivFD~\cite{UFD}, NPR~\cite{NPR}, RINE~\cite{RINE}, FatFor.~\cite{fatformer}, DRCT~\cite{DRCT}, PatchFor.~\cite{PatchFor}, and FSD~\cite{fsd}) \& two attribution methods (RepMix~\cite{Repmix} and POSE~\cite{POSE}).

\subheader{Implementation Details}
We train the projection network $f$ using AdamW~\cite{AdamW} (\(5 \times 10^{-5}\) learning rate, \(0.01\) weight decay) with mini-batches of 256 samples and 128 hardest positives/negatives per anchor for 10 epochs. The distribution of embeddings are modeled using GMMs~\cite{GMM} with a confidence threshold set to maximize the difference between true rejection rate and false rejection rate. Additional implementation details are in the supplementary material Sec.~A.

\subsection{Sequentially Adapting to Multiple New Sources}
\label{sec:main_results}

\subheader{Setup} We evaluated our system's ability to detect synthetic images as new generators continuously emerge, adhering to the following protocol:
The system is initialized with a set of real images and known AI-generated images.
We simulate regular operation by feeding the system an unlabeled stream of images. This stream contains real images, synthetic images from known generators, and from a sequence of emerging unknown generators introduced one at each stage. The stream maintains an equal ratio for all sources (real and each AI generator).

We note that all methods are only initialized on the same training dataset. Additionally, to ensure our results are independent of emergence order, we report the average performance over 10 randomized orderings of the new generators.

\subheader{Metrics}
We report each stage's balanced detection accuracy (\%) including all previous generators,
 and the drop between the initial and final accuracy.

\subheader{Results}
The results in Tab.~\ref{tab:main_results} show that AS-AID sustains high performance as new generators continuously
emerge. Our system's final accuracies reach 94.5\% on DSD and 95.4\% on GID, exceeding most competitors by more than 20 percentage points at the last stage. The only close baseline is FSD, which AS-AID surpasses by 6\% on DSD and 8\% on GID.
AS-AID achieves these gains despite having a lower initial accuracy, underscoring its adaptive capability.

In contrast, most competing methods suffer significant performance drops. The majority experienced a performance drop of 20 percentage points or greater. For example, methods with near-perfect initial accuracy on DSD (RINE: 100.0\%, POSE: 99.8\%) still suffered severe drops of 25.5 and 39.0 percentage points, respectively, revealing that high initial performance does not guarantee overall performance. A notable exception is DRCT, which shows a low total drop similar to ours. However, this is an artifact of its low initial accuracy (74.6\%), and its final performance remained over 20 percentage points lower than AS-AID on both datasets, with final accuracies of only 72.1\% and 71.5\%.

\subsection{Adapting to Individual New Sources}
\label{sec:per_source_analysis}

\subheader{Setup}
This experiment analyzes the system's performance variability when adapting to individual new sources. This is designed to measure performance consistency, as some generative sources may be easily detected via simple transferability, while others are more difficult and require true adaptation. To do this, we initialized our system as in Sec.~\ref{sec:main_results}, introduced a single emerging source, let it adapt, then measured its detection performance. We repeated this for all seven emerging sources in the GenImage dataset and aggregated the results.

\subheader{Metrics}
We report the mean and standard deviation of the balanced detection accuracy across all generators, as well as the worst-case generator accuracy.

\begin{table}[t]
	\centering
	\scriptsize
	\caption{Performance analysis on individual sources in GID.}
	\label{tab:individual_perf}
	\setlength{\tabcolsep}{3pt}
	\begin{tabular}{l c c c c @{\hspace{1.2ex}} c}
		\toprule
		\textbf{Method} & \textbf{Known} & \textbf{New Avg.} & \textbf{Std.} & \multicolumn{2}{c}{\textbf{Worst}} \\
		\midrule
		CCNSpot   & 76.7 & 56.6 & 9.4 & 49.7 & (BigG.) \\
		NPR       & 93.3 & 75.2 & 14.9 & 51.4 & (BigG.) \\
		RINE      & \textbf{100.0} & 74.0 & 19.3 & 50.4 & (BigG.) \\
		POSE      & 99.8 & 71.7 & 18.4 & 52.8 & (ADM) \\
		FSD       & 98.5 & 81.9 & 21.1 & 48.5 & (BigG.) \\
		FatFormer & 98.4 & 67.2 & 17.6 & 50.0 & (BigG.) \\
		UnivFD    & 85.6 & 71.4 & 10.4 & 55.0 & (ADM) \\
		DRCT      & 74.6 & 71.1 & 2.7 & 66.1 & (ADM) \\
		RepMix    & 96.0 & 65.8 & 18.6 & 49.4 & (BigG.) \\
		\midrule
		\textbf{Ours} & 97.5 & \textbf{97.2} & \textbf{1.4} & \textbf{94.8} & (Midj.) \\
		\bottomrule
	\end{tabular}
	\vspace{-1.5em}
\end{table}

\subheader{Results}
The results in Tab.~\ref{tab:individual_perf} show that AS-AID maintains strong, consistent detection performance across all new generators.
It achieves a high mean accuracy of 97.2\%, a low standard deviation of 1.4, and a worst-case accuracy of 94.8\%. AS‑AID achieves this by leveraging its adaptive forensic embedding space and new source discovery module to integrate emerging forensic traces.

By contrast, the best competitor, FSD, only achieves an average accuracy of 81.9\% on detecting images from new sources. We also observe that all other methods, including FSD, have low worst-case performance. This is because each of them are weak against at least one generators due to these sources having significantly different forensic microstructures than those learned during training.

\subsection{Comparison with Naive Self-Adaptation}
\label{sec:principled_design}

\begin{figure}[t]
	\vspace{-1em}
	\centering
	\includegraphics[width=0.8\linewidth]{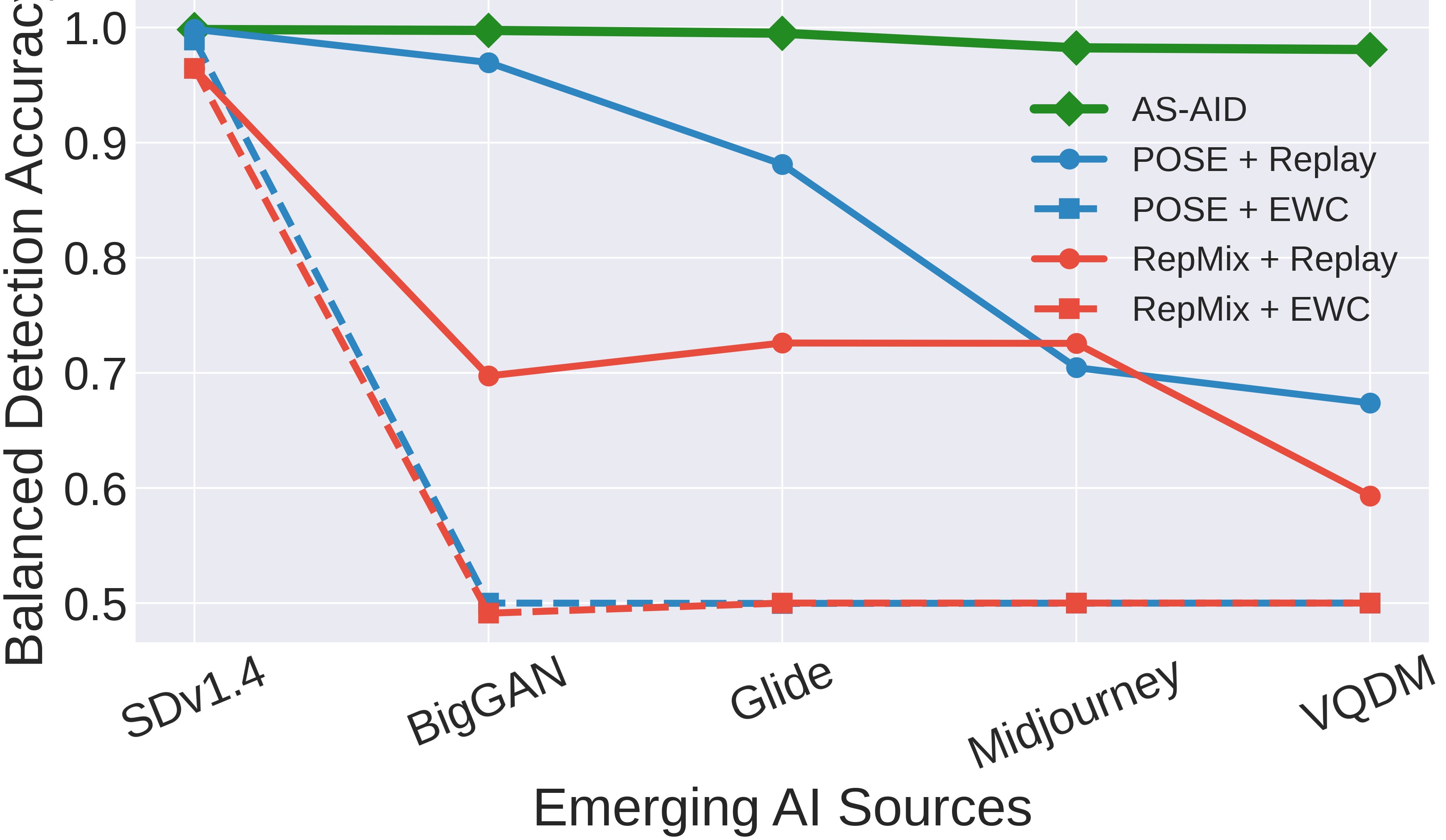}
	\caption{Comparison with naive self-adaptive methods.}
	\label{fig:adaptation_results} 
	\vspace{-1.5em}
\end{figure}

\subheader{Setup}
In this experiment, we compared  AS-AID to  naively constructed self-adaptive systems.
While AS-AID is the first proposed autonomous self-adaptive system for synthetic image detection,  other self-adaptive systems can be naively
constructed by combining an open set source attribution system with a continual learning method.  Here, all images with unknown sources are grouped together to form training data for a new source.  The continual learning method then uses this data to update the open-set attributer.

We use two open-set methods, POSE~\cite{POSE}, RepMix~\cite{Repmix}, combined with two continual learning approaches, data replay~\cite{icarl}, Elastic Weight Consolidation~\cite{ewc}. We evaluate these as four generative sources are sequentially introduced.

\subheader{Metrics}
Average balanced detection accuracy (\%) after introducing each source.

\subheader{Results}
Fig.~\ref{fig:adaptation_results} shows that all naive systems degrade rapidly after only a few new generators. Methods updated with the continual learning approach Elastic Weight Consolidation collapse after a single update, losing discrimination between real and synthetic images. Data replay is more resilient but still suffers a sharp drop within a few cycles.

The primary cause is data contamination. As we will quantify in Sec.~\ref{sec:discovery_performance}, the buffer of naive systems contains many mislabeled instances from known sources. Updating on this noisy data progressively corrupts their internal models~\cite{coteaching, Mentornet, understanding}.
In contrast, AS‑AID's new source discovery and autonomous validation subsystems provide the safeguards essential for long‑term performance reliability, ensuring consistent accuracy throughout the experiment.

\subsection{Effects of Data Purity for Reliable Adaptation}
\label{sec:discovery_performance}

\subheader{Setup}
To understand why naive approaches to self-adaptation suffer rapid performance declines, we conducted an experiment where we analyzed the data that both AS-AID and naive self-adaptive approaches used to model a new generator.
To do this,
we sequentially introduced new generators from DSD and examined the composition of the data that each approach attributed to a new generator.

\subheader{Metrics}
We report two metrics: Discovery Purity, the proportion of samples in a discovered cluster that truly originate from the new source, and Discovery Coverage, the proportion of all new-source samples captured by that cluster. Both metrics are computed prior to any model adaptation.

\begin{table}[t]
	\centering
	\scriptsize
	\caption{Source discovery performance on DSD.}
	\label{tab:discovery_performance}

	\setlength{\tabcolsep}{3pt}
	\resizebox{0.75\columnwidth}{!}{
		\begin{tabular}{l c c c}
			\toprule
			\textbf{Metric (\%)} & \textbf{Ours}           & POSE~\cite{POSE} & RepMix~\cite{Repmix}    \\
			\midrule
			Discovery Purity     & \textbf{98.1 $\pm$ 3.2} & 75.1 $\pm$ 31.2  & 31.3 $\pm$ 16.6         \\
			Discovery Coverage   & 23.0 $\pm$ 6.3          & 55.6 $\pm$ 37.7  & \textbf{88.8 $\pm$ 2.3} \\
			\bottomrule
		\end{tabular}
	}
\end{table}

\begin{figure}[t]
	\centering
	\begin{subfigure}[b]{0.3\linewidth}
		\centering
		\includegraphics[width=\linewidth]{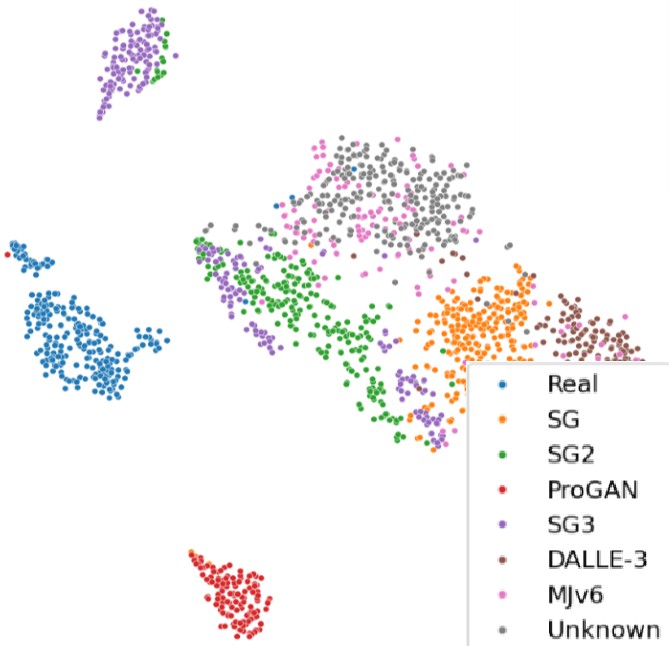}
	\end{subfigure}%
	\hfill\vrule\hfill
	\begin{subfigure}[b]{0.3\linewidth}
		\centering
		\includegraphics[width=\linewidth]{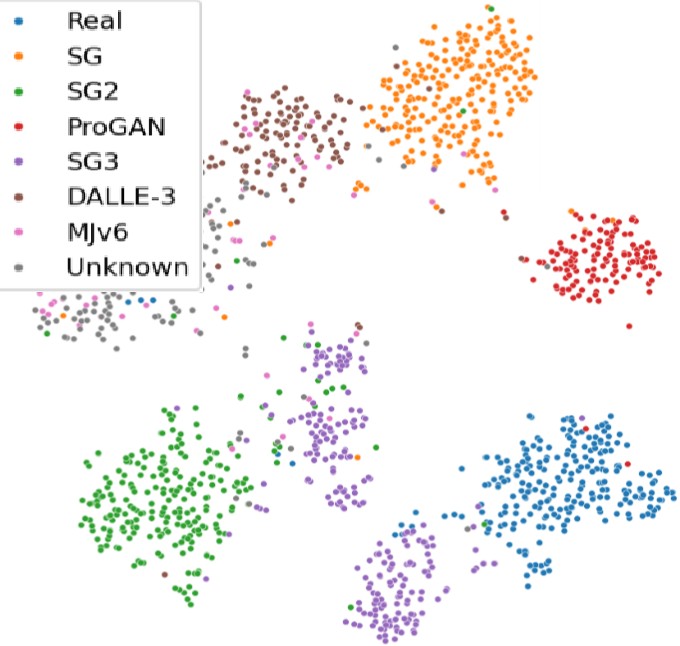}
	\end{subfigure}%
	\hfill\vrule\hfill
	\begin{subfigure}[b]{0.3\linewidth}
		\centering
		\includegraphics[width=\linewidth]{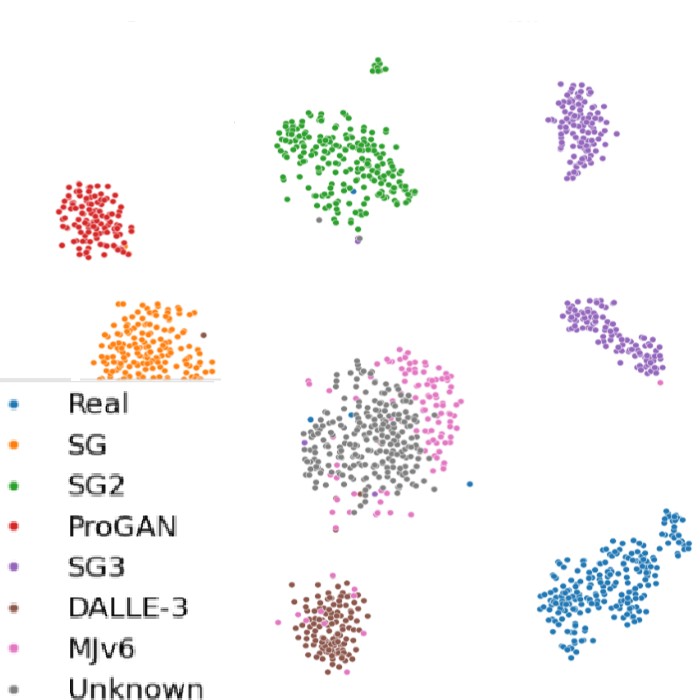}
	\end{subfigure}
	\caption{t-SNE embedding space plots of original (left), enhanced source separation (middle), enhanced source separation + preservation (right).}
	\label{fig:enhance_sep_space_loss}
	\vspace{-1.5em}
\end{figure}

\subheader{Results}
The results in Tab.~\ref{tab:discovery_performance} show our AS-AID system
achieves a high Discovery Purity of 98.1\%, demonstrating that our discovery algorithm correctly identifies data from new generative sources.
This prevents AS-AID from suffering performance declines due to data contamination~\cite{Mentornet, coteaching, understanding}.

Our system intentionally prioritizes discovery purity over coverage, since updating the model using low-purity data risks degrading detection performance across all generators, whereas delaying adaptation still permits detection through transferability until sufficient high-purity evidence of a new source is obtained. In contrast, other methods achieve higher Coverage but suffer significant Purity drops, as they treat all unknown-source images as originating from a new generator, leading to substantial degradation from data contamination. 

\section{Discussion}
\label{sec:data_efficiency}

\subheader{AS-AID Data Requirement Analysis}
Here, we examine the number of images required for successful discovery of the emergence of a new source. To do this, we start with having $Y=100$ images from a new source and evaluate the system's ability to discover this source. We repeat this experiments with at an increment of 100 until $Y=1000$ and report the results in Tab.~\ref{tab:data_efficiency}.

\begin{table}[t]
	\centering
	\scriptsize
	\caption{Discovery rate vs observed image count.}
	\label{tab:data_efficiency}
	\setlength{\tabcolsep}{3pt}
	\begin{tabular}{l c c c c c}
		\toprule
		\textbf{\# Images Introduced} & 100  & 300  & 500  & 750  & 1000 \\
		\midrule
		Discovery Rate (\%)           & 22.9 & 68.6 & 85.7 & 85.7 & 95.7 \\
		\bottomrule
	\end{tabular}
\end{table}

From the results, we see that our AS-AID system demonstrates high data efficiency: the discovery rate, while relatively low at 100 samples (22.9\%), increases to 85.7\% by 500 samples and reaches 95.7\% with 1000 samples. Importantly, this efficiency does not compromise the high purity established in Sec.~\ref{sec:discovery_performance}, confirming that the system achieves reliable discovery with modest data requirements.

\begin{table}[t]
	\centering
	\scriptsize
	\caption{Embedding space vs. performance.}
	\label{tab:three_versions}
	\begin{tabular}{l c c}
		\toprule
		\textbf{Embed Space}   & \textbf{Det. Acc}     & \textbf{Attr. Acc}    \\
		\midrule
		FSD               & 85.8          & 75.1          \\
		FSD + Separation  & 94.3          & 81.6          \\
		\midrule
		\textbf{Proposed} & \textbf{99.2} & \textbf{89.5} \\
		\bottomrule
	\end{tabular}
\end{table}

\subheader{Analysis of Embedding Space Choice}
We evaluate how different embedding spaces affect detection and attribution: (1) the original FSD space, (2) FSD with separation loss (Eq.~\ref{eq:sep_loss}), and (3) our Enhanced Forensic Embedding Space.
As shown in Tab.~\ref{tab:three_versions} and Fig.~\ref{fig:enhance_sep_space_loss}, the original FSD space establishes a baseline, but adding separation loss improves performance by 8.5\% and 6.5\% for detection and attribution, respectively. Our proposed loss yields the best embedding space, achieving 99.2\% detection and 89.5\% attribution accuracy.

\begin{figure}[t]
	\centering
	\begin{subfigure}[b]{0.22\linewidth}
		\centering
		\includegraphics[width=\linewidth]{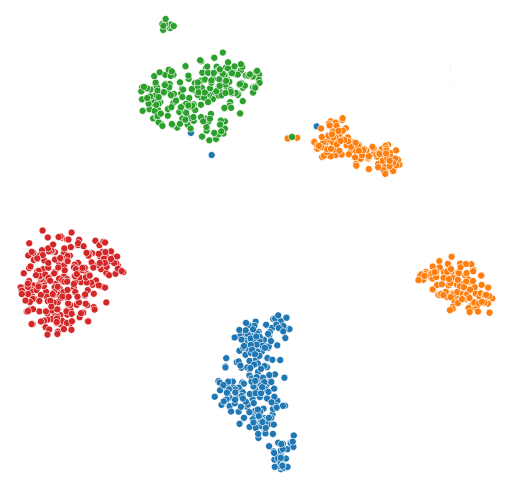}
	\end{subfigure}%
	\hfill\vrule\hfill
	\begin{subfigure}[b]{0.22\linewidth}
		\centering
		\includegraphics[width=\linewidth]{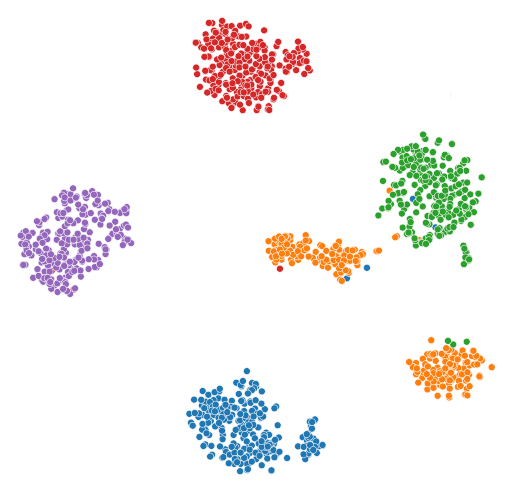}
	\end{subfigure}%
	\hfill\vrule\hfill
	\begin{subfigure}[b]{0.22\linewidth}
		\centering
		\includegraphics[width=\linewidth]{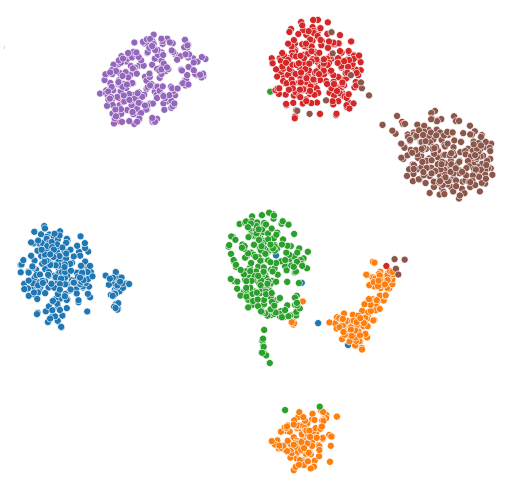}
	\end{subfigure}%
	\hfill\vrule\hfill
	\begin{subfigure}[b]{0.22\linewidth}
		\centering
		\includegraphics[width=\linewidth]{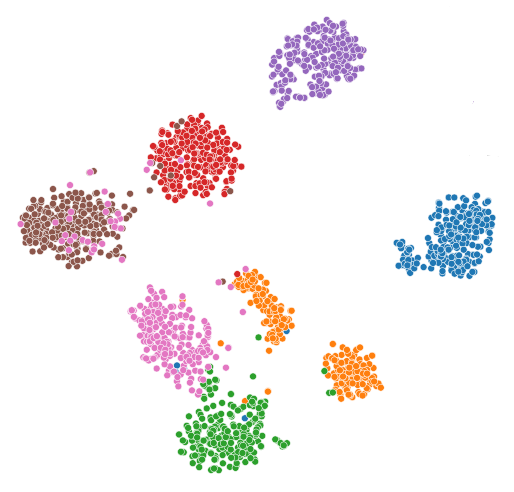}
	\end{subfigure}
	\caption{t-SNE of evolving embedding space at time step 0, 1, 2, and 3 (left to right).}
	\label{fig:embed_space_over_time}
	\vspace{-1.0em}
\end{figure}

\subheader{Continuously Evolving Embedding Space}
Fig.~\ref{fig:embed_space_over_time} illustrates the adaptive capability of our framework. The sequence of t-SNE plots visualizes the progressive evolution of the forensic embedding space as new sources are introduced. At each stage, the system incorporates the new generator by forming a new, distinct, and well-separated cluster. This behavior contrasts with the fixed embedding spaces of existing paradigms, which are unable to reliably capture the novel forensic signatures of emerging sources.

\subheader{Limitations \& Future Work}
Our system may sometimes miss the emergence of a new source if it is too similar to a known one. Despite this, overall synthetic image detection remains strong because the system associates these with generative sources. Furthermore, multiple generators may emerge simultaneously, leading to sub‑optimal performance for new source discovery. Understanding and mitigating these risks are important directions for future work.

\begin{table}[t]
	\centering
	\scriptsize
	\caption{Ablation study of our proposed framework.}
	\vspace{-1em}
	\setlength{\tabcolsep}{1pt} 
	\label{tab:ablation_study}
	\resizebox{\linewidth}{!}{%
		\begin{tabular}{l c ccc ccc ccc ccc}
			\toprule
			& \textbf{Proposed} & \multicolumn{3}{c}{Clustering Alg.} & \multicolumn{3}{c}{Embedding Space} & \multicolumn{3}{c}{Discovery Criteria} & \multicolumn{3}{c}{Validation} \\
			\cmidrule(lr){2-2} \cmidrule(lr){3-5} \cmidrule(lr){6-8} \cmidrule(lr){9-11} \cmidrule(lr){12-14}
			\textbf{Metric} & Best & None & k-means & HDB. & No Pres. & Imbal. & CLIP & None & Suff. & Cohes. & None & Det. & Qual. \\
			\midrule
			Det. Acc  & \textbf{99.2} & 90.6 & 94.5 & 95.4 & 94.3 & 93.7 & 72.5 & 96.4 & 98.7 & 93.6 & 98.4 & 98.7 & 97.3 \\
			Attr. Acc & \textbf{89.5} & 40.7 & 50.1 & 61.4 & 81.6 & 69.0 & 34.6 & 66.4 & 65.5 & 76.3 & 75.2 & 68.2 & 61.4 \\
			\bottomrule
		\end{tabular}%
	}
	\vspace{-1.5em}
\end{table}

\section{Ablation}
\label{sec:ablation}

We conducted an ablation to assess the impact of our design choices by adopting the setup of experiment in Sec.~\ref{sec:principled_design}. Results in Tab.~\ref{tab:ablation_study} show that each of our proposed subsystem is critical for maintaining strong performance.

\subheader{Clustering Algorithm}
We tested the New Source Discovery Algorithm (Sec.~\ref{sec:discovery}) with no clustering, with k-means, and with HDBSCAN~\cite{hdbscan}. Removing clustering or using k-means~\cite{kmeans} harms attribution and weakens detection. While HDBSCAN improves performance, it falls short of our proposed system using DBSCAN.

\subheader{Embedding Space}
We replaced Enhanced Forensic Embedding Space with alternatives. CLIP embeddings fail for source attribution and detection. Removing the preservation term or using an imbalanced loss also degrades performance, confirming the necessity of a balanced separation-preservation loss ($\lambda=1$).

\subheader{Discovery Criteria}
We ablated the sufficiency and cohesion criteria in our New Source Discovery Algorithm (Sec.~\ref{sec:discovery}). Dropping both causes large performance loss. Removing only one criterion also severely harms attribution, showing that both are complementary and required for reliable discovery.

\subheader{Validation}
We removed validation checks in the adaptation subsystem (Sec.~\ref{sec:validation}). Eliminating validation altogether reduces performance. Removing only the detection performance check or only the model quality check leads to performance decline, showing that both safeguards are essential for reliable performance.

\section{Conclusion}
We introduced AS-AID, an autonomous framework for AI-generated image identification that adapts to emerging sources without human supervision. By integrating identification, discovery, and self-adaptation into a unified system, AS-AID maintains strong detection performance as new generators emerge while existing methods degrade. Our results demonstrate that autonomous adaptation can substantially improve performance in evolving generative environments.

\bibliographystyle{splncs04}
\bibliography{main}

\end{document}